\documentclass[letterpaper]{article} 
\usepackage[draft]{aaai2026}
\usepackage{times}  
\usepackage{helvet}  
\usepackage{courier}  
\usepackage[hyphens]{url}  
\usepackage{graphicx} 
\usepackage{natbib}  
\usepackage{caption} 
\usepackage{algorithm}
\usepackage{algorithmic}

\usepackage{newfloat}
\usepackage{listings}
\DeclareCaptionStyle{ruled}{labelfont=normalfont,labelsep=colon,strut=off} 
\floatstyle{ruled}
\newfloat{listing}{tb}{lst}{}
\floatname{listing}{Listing}
\title{A Multi-Agent Feedback System for Detecting and Describing News Events in Satellite Imagery}
\author {
    Madeline Anderson\textsuperscript{\rm 1},
    Mikhail Klassen\textsuperscript{\rm 2},
    Ash Hoover\textsuperscript{\rm 2},
    Kerri Cahoy\textsuperscript{\rm 1}
}
\affiliations {
    \textsuperscript{\rm 1}Massachusetts Institute of Technology,
    \textsuperscript{\rm 2}Planet Labs
}

\usepackage{bibentry}

\begin{document}

\maketitle

\begin{abstract}


Changes in satellite imagery often occur over multiple time steps. Despite the emergence of bi-temporal change captioning datasets, there is a lack of multi-temporal event captioning datasets (at least two images per sequence) in remote sensing. This gap exists because (1) searching for visible events in satellite imagery and (2) labeling multi-temporal sequences require significant time and labor. To address these challenges, we present SkyScraper, an iterative multi-agent workflow that geocodes news articles and synthesizes captions for corresponding satellite image sequences. Our experiments show that SkyScraper successfully finds 5x more events than traditional geocoding methods, demonstrating that agentic feedback is an effective strategy for surfacing new multi-temporal events in satellite imagery. We apply our framework to a large database of global news articles, curating a new multi-temporal captioning dataset with 5,000 sequences. By automatically identifying imagery related to news events, our work also supports journalism and reporting efforts.
\end{abstract}

\section{Introduction}
Traditional remote sensing change captioning methods typically rely on manual annotation or rules-based approaches applied to existing datasets with fine-grained change labels (e.g., segmentation maps). Because labels are scarce, these methods usually operate on bi-temporal image pairs focusing on land-use land-cover (LULC) changes, such as building and road modifications \cite{liu2022remote,hoxha2022change,karaca2025robust,yuan2022change}. Recent work incorporates Large Language Models (LLMs) to improve scalability and caption diversity \cite{deng2025changechat,noman2024cdchat,irvin2024teochat,elgendy2024geollava}. However, these approaches still rely on pre-labeled temporal datasets and inherit similar limitations. Multi-temporal captioning datasets (with more than two images) remain largely restricted to UAV video \cite{bashmal2023capera}.

Expanding current datasets remains difficult because surfacing \textit{new} change events in satellite imagery and annotating multi-temporal image sequences demand significant time and labor. Building on recent work that curates a multi-temporal dataset by geocoding FEMA-sourced disaster reports \cite{revankar2025monitrs}, we generalize the approach to global news articles and introduce multi-agent feedback for event geocoding and captioning, producing a new multi-temporal captioning dataset spanning diverse global events.


\section{Traditional Geocoding}


For rules-based geocoding methods, we first extract all named geographic entities from the article, each with associated geocoded coordinates $c_i=(\phi_i, \lambda_i)$ and weight $w_i$, which is the number of times that name appears in the article. We experiment with two traditional geocoding methods: (1) weighted centroid, and (2) Geo-referenced Information Processing SYstem (GIPSY) \cite{woodruff1994gipsy}. The weighted centroid approach converts the coordinates to cartesian Earth-Centered, Earth-Fixed (ECEF) format, then computes their weighted average. Alternatively, GIPSY is a weighted polygon-based approach that represents each candidate location as a 3D polyhedron whose base corresponds to its geographic bounding box, 
and whose height, $z_i$, corresponds with its weight $w_i$. 
It then overlays the polyhedra based on the following criteria:
\begin{enumerate}
    \item No overlap: place at $z=0$.
    \item Fully contained: stack on top of the containing polygon.
    \item Partial overlap: place non-overlapping part at $z=0$, stack overlapping parts on intersecting regions.
\end{enumerate}
After adding all polygons, GIPSY estimates the article location using the centroid of the highest-elevation region.

\section{Approach}
\begin{figure}[!htb]
    \centering
    \includegraphics[width=0.98\linewidth]{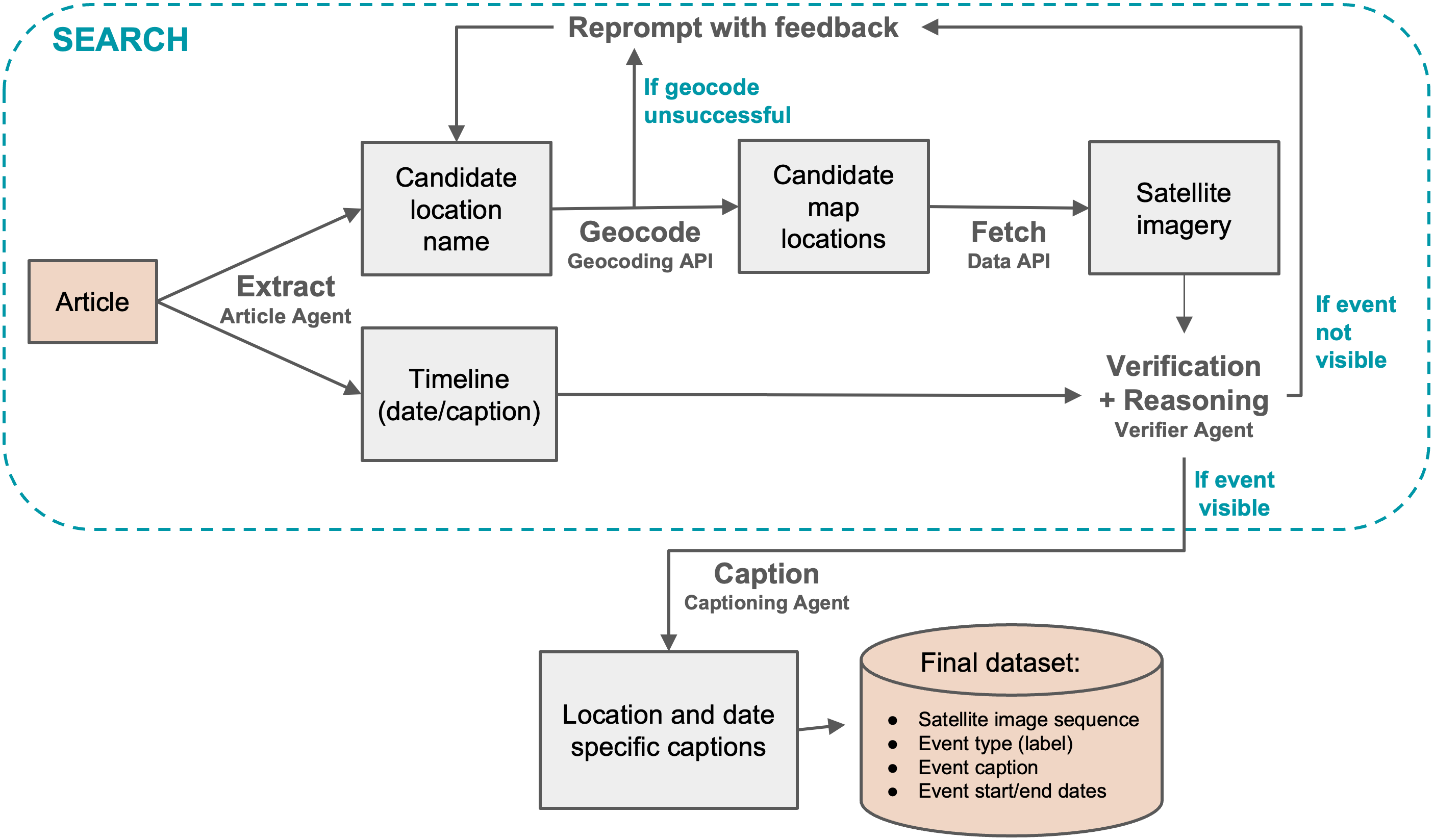}
    \caption{SkyScraper iterative feedback pipeline.}
    \label{fig:pipeline}
\end{figure}

Traditional geocoding methods are susceptible to noise, such as incorrect location names and geocoding inaccuracies. Moreover, the frequency that an article mentions a location (used as the weighting factor) does not always correlate with event occurrence \textit{and} visibility at that location. 
SkyScraper addresses these limitations using an agentic iterative approach (Figure \ref{fig:pipeline}) with the following steps:
\begin{enumerate}
    \item \textbf{Extract:} An LLM \textit{article agent} extracts a named geographic entity and event timeline from the article text.
    \item \textbf{Geocode:} A geocoding API converts named geographic entities to latitude-longitude coordinates.
    \item \textbf{Fetch:} A data API retrieves satellite imagery at the geocoded coordinates over the timeline range.
    \item \textbf{Verify:} A multimodal LLM \textit{verifier agent} cross-references the article and fetched imagery to verify whether the event is visible while providing reasoning. 
    \item \textbf{Caption:} A multimodal LLM \textit{captioning agent} writes a change caption, using the article as context.
\end{enumerate}

Rather than extracting all location names at once, the system requests one candidate location at a time and repeats steps 1-4 to refine searches if (1) geocoding fails or (2) the verifier agent does not detect the event. In each case, it requests a new candidate, incorporating the failed location and reasoning (Algorithm \ref{alg:feedback}). If it does not find the event before reaching the maximum attempts, it returns an empty result.

\begin{algorithm}[!htb]
\fontsize{8.5}{10.5}\selectfont
\begin{flushleft}
\caption{Iterative Search and Verification}
\label{alg:feedback}
\begin{algorithmic}[1]
\REQUIRE Article text $T$, maximum attempts \textit{MAX\_ATTEMPTS}

\STATE $attempts \leftarrow 0$
\STATE $failures \leftarrow$ empty list
\STATE $\ell^* \leftarrow$ \texttt{None}

\WHILE{$attempts <$ \textit{MAX\_ATTEMPTS}}
    \STATE $attempts \leftarrow attempts + 1$
    \STATE $\ell_t \leftarrow$ ArticleAgent($T$, $failures$)
    \STATE $c_t \leftarrow$ GeocodeAPI($\ell_t$)
    
    \IF{$c_t$ is invalid}
        \STATE Append($failures$, ($\ell_t$, \texttt{geocode\_error}))
        \STATE \textbf{continue}
    \ELSE
    
    \STATE $D_t \leftarrow$ DataAPI($c_t$)
    \STATE $(visible, reason) \leftarrow$ VerifierAgent($T$, $D_t$)
    
    \IF{$visible = \texttt{false}$}
        \STATE Append($failures$, ($\ell_t$, \texttt{reason}))
        \STATE \textbf{continue}
    \ELSE
        \STATE $\ell^* \leftarrow \ell_t$
        \STATE CaptioningAgent($T$, $D_t$)
        \STATE \textbf{break}
    \ENDIF
    \ENDIF
\ENDWHILE
\end{algorithmic}
\end{flushleft}
\end{algorithm}
\section{Experiments}
We collected 1,000 news articles using the Google Search API, covering diverse global events such as natural disasters, construction, and civil unrest. We applied each geocoding approach (weighted centroid, GIPSY, and SkyScraper agentic feedback) to retrieve PlanetScope imagery (3m resolution). We used Mapbox as the geocoding API and Gemini-2.5-flash for the LLM agents. 

We manually verified positive event detections for each approach. Table \ref{tab:geocoding_results} reports the geocoding performance, where \textit{yield} represents the percentage of correct detections (true positives) out of the initial set of articles. Because we only count true positives, articles without visible events reduce raw yield, making relative improvement over the centroid baseline more informative than absolute yield.

\begin{table}[!htbp]
\centering
\small
\begin{tabular}{lccc}
\textbf{Method} 
& \textbf{\shortstack{Event\\Detections}} 
& \textbf{Yield} 
& \textbf{\shortstack{Increase over\\Centroid Baseline}} \\
\hline \\[-1.5ex]
Weighted Centroid  & 17 & 1.7\% & -- \\
GIPSY              & 47 & 4.7\% & 2.8$\times$ \\
SkyScraper (ours)  & 84 & 8.4\% & 4.9$\times$ \\
\end{tabular}
\caption{Event detection yield and relative improvement over centroid baseline for news article geocoding.}
\label{tab:geocoding_results}
\end{table}




\section{Dataset Curation}

\begin{figure}[!htb]
    \centering
    \includegraphics[width=\linewidth]{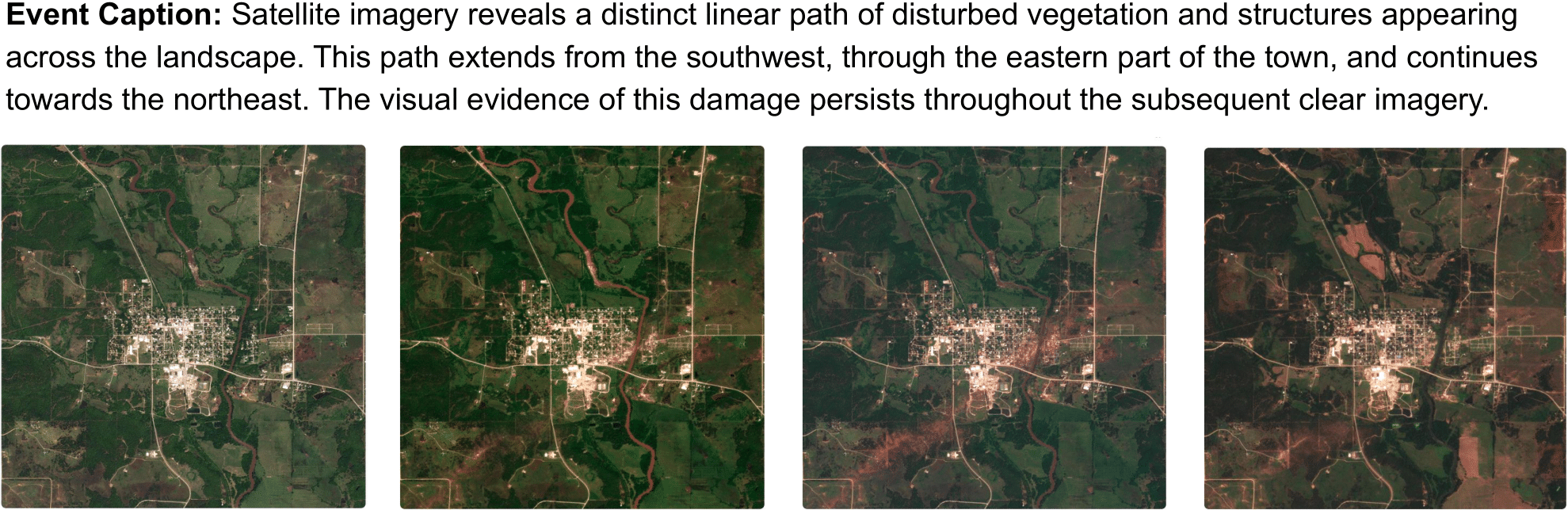}
    \caption{PlanetScope imagery and caption for a geocoded tornado event identified by SkyScraper.}
    \label{fig:example}
\end{figure}

We applied SkyScraper to news articles sampled from the Global Database of Events, Language, and Tone\footnote{\url{https://www.gdeltproject.org/}} (GDELT) \cite{leetaru2013gdelt} from 2022–2024 using PlanetScope and Sentinel-2 imagery. Annotators verified captions and event dates to produce the final datasets (Table \ref{tab:dataset_stats}). Figure \ref{fig:example} shows a detected tornado event. The dataset is available at \url{https://source.coop/planet/skyscraper}.


\begin{table}[!htbp]
\centering
\small
\setlength{\tabcolsep}{5pt}
\begin{tabular}{lccc}
\textbf{\shortstack[t]{\smash{Imagery}\\Source}} 
& \textbf{\shortstack[t]{Total\\Sequences}} 
& \textbf{\shortstack[t]{Confirmed\\Events}} 
& \textbf{\shortstack[t]{\smash{Average} \# \smash{Images}\\Per Sequence}} \\
\hline \\[-1.5ex]
PlanetScope  & 5,000 & 3,288 & 21 \\
Sentinel-2   & 4,463 & 2,931 & 8 \\
\end{tabular}
\caption{SkyScraper GDELT dataset statistics.}
\label{tab:dataset_stats}
\end{table}


\section{Conclusion}


We introduce SkyScraper, a novel multi-agent feedback system that locates and captions events in multi-temporal satellite imagery using news articles. Compared to traditional geocoding methods, our approach increases event detections by nearly 5x. We apply SkyScraper to the GDELT database to produce new PlanetScope and Sentinel-2 multi-temporal captioning datasets with $\sim$5,000 sequences. These results demonstrate the effectiveness of agentic feedback for facilitating event geocoding, multi-image captioning, and benchmark dataset curation in the remote sensing domain.
\section{Acknowledgments}
We would like to thank Planet Labs for providing funding support. We are also grateful to MIT students Kunal Rajadhyax, Bobo Lin, Benjamin Gao, Lindy Zhang, and Dustin Wang for assisting with manual dataset annotation.

\bibliography{aaai2026}

@article{liu2022remote,
  title={Remote sensing image change captioning with dual-branch transformers: A new method and a large scale dataset},
  author={Liu, Chenyang and Zhao, Rui and Chen, Hao and Zou, Zhengxia and Shi, Zhenwei},
  journal={IEEE Transactions on Geoscience and Remote Sensing},
  volume={60},
  pages={1--20},
  year={2022},
  publisher={IEEE}
}

@article{hoxha2022change,
  title={Change captioning: A new paradigm for multitemporal remote sensing image analysis},
  author={Hoxha, Genc and Chouaf, Seloua and Melgani, Farid and Smara, Youcef},
  journal={IEEE Transactions on Geoscience and Remote Sensing},
  volume={60},
  pages={1--14},
  year={2022},
  publisher={IEEE}
}

@article{karaca2025robust,
  title={Robust change captioning in remote sensing: Second-cc dataset and mmodalcc framework},
  author={Karaca, Ali Can and Ozelbas, Enes and Berber, Saadettin and Karimli, Orkhan and Yildirim, Turabi and Amasyali, M Fatih},
  journal={IEEE Journal of Selected Topics in Applied Earth Observations and Remote Sensing},
  year={2025},
  publisher={IEEE}
}

@article{yuan2022change,
  title={Change detection meets visual question answering},
  author={Yuan, Zhenghang and Mou, Lichao and Xiong, Zhitong and Zhu, Xiao Xiang},
  journal={IEEE Transactions on Geoscience and Remote Sensing},
  volume={60},
  pages={1--13},
  year={2022},
  publisher={IEEE}
}

@inproceedings{deng2025changechat,
  title={Changechat: An interactive model for remote sensing change analysis via multimodal instruction tuning},
  author={Deng, Pei and Zhou, Wenqian and Wu, Hanlin},
  booktitle={ICASSP 2025-2025 IEEE International Conference on Acoustics, Speech and Signal Processing (ICASSP)},
  pages={1--5},
  year={2025},
  organization={IEEE}
}

@article{noman2024cdchat,
  title={Cdchat: A large multimodal model for remote sensing change description},
  author={Noman, Mubashir and Ahsan, Noor and Naseer, Muzammal and Cholakkal, Hisham and Anwer, Rao Muhammad and Khan, Salman and Khan, Fahad Shahbaz},
  journal={arXiv preprint arXiv:2409.16261},
  year={2024}
}

@article{irvin2024teochat,
  title={Teochat: A large vision-language assistant for temporal earth observation data},
  author={Irvin, Jeremy Andrew and Liu, Emily Ruoyu and Chen, Joyce Chuyi and Dormoy, Ines and Kim, Jinyoung and Khanna, Samar and Zheng, Zhuo and Ermon, Stefano},
  journal={arXiv preprint arXiv:2410.06234},
  year={2024}
}

@article{elgendy2024geollava,
  title={Geollava: Efficient fine-tuned vision-language models for temporal change detection in remote sensing},
  author={Elgendy, Hosam and Sharshar, Ahmed and Aboeitta, Ahmed and Ashraf, Yasser and Guizani, Mohsen},
  journal={arXiv preprint arXiv:2410.19552},
  year={2024}
}

@article{bashmal2023capera,
  title={Capera: Captioning events in aerial videos},
  author={Bashmal, Laila and Bazi, Yakoub and Al Rahhal, Mohamad Mahmoud and Zuair, Mansour and Melgani, Farid},
  journal={Remote Sensing},
  volume={15},
  number={8},
  pages={2139},
  year={2023},
  publisher={MDPI}
}

@article{revankar2025monitrs,
  title={MONITRS: Multimodal Observations of Natural Incidents Through Remote Sensing},
  author={Revankar, Shreelekha and Mall, Utkarsh and Phoo, Cheng Perng and Bala, Kavita and Hariharan, Bharath},
  journal={arXiv preprint arXiv:2507.16228},
  year={2025}
}

@misc{woodruff1994gipsy,
  title={GIPSY: Georeferenced Information},
  author={Woodruff, Allison G and Plaunt, Christian},
  year={1994},
  publisher={University of California at Berkeley}
}

@inproceedings{leetaru2013gdelt,
  title={Gdelt: Global data on events, location, and tone, 1979--2012},
  author={Leetaru, Kalev and Schrodt, Philip A},
  booktitle={ISA annual convention},
  volume={2},
  number={4},
  pages={1--49},
  year={2013},
  organization={Citeseer}
}

\end{document}